# Material Editing Using a Physically Based Rendering Network

Guilin Liu[1,3*]  Duygu Ceylan[2]  Ersin Yumer[2]  Jimei Yang[2]  Jyh-Ming Lien[1]
[1]George Mason University  [2]Adobe Research  [3]NVIDIA

## Abstract

*The ability to edit materials of objects in images is desirable by many content creators. However, this is an extremely challenging task as it requires to disentangle intrinsic physical properties of an image. We propose an end-to-end network architecture that replicates the forward image formation process to accomplish this task. Specifically, given a single image, the network first predicts intrinsic properties, i.e. shape, illumination, and material, which are then provided to a rendering layer. This layer performs in-network image synthesis, thereby enabling the network to understand the physics behind the image formation process. The proposed rendering layer is fully differentiable, supports both diffuse and specular materials, and thus can be applicable in a variety of problem settings. We demonstrate a rich set of visually plausible material editing examples and provide an extensive comparative study.*

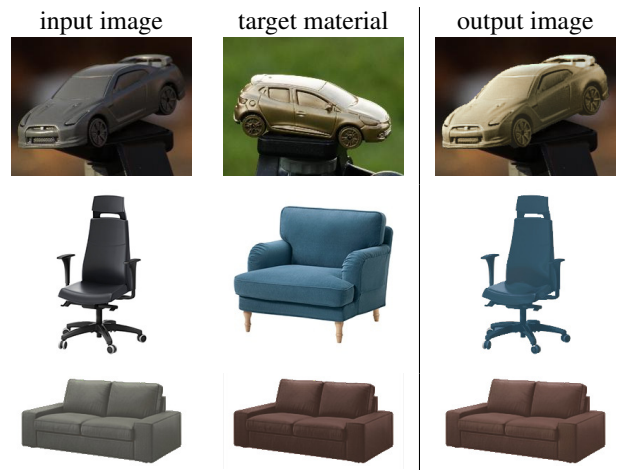

Figure 1. We replace the materials of the objects in the input images with the material properties of the objects in the middle column, resulting edited images are shown in the right column.

## 1. Introduction

One of the main properties of an object that contributes to its appearance, is material. Hence, many designers desire to effortlessly alter materials of objects. In case of 2D design, however, often the designer only has access to a single image of the object. Given one such image, e.g. an image of a fabric sofa, how would you synthesize a new image that depicts the same sofa with a more specular material?

A typical approach to addressing this image-based material editing problem is to first infer the intrinsic properties (e.g. shape, material, illumination) from the image which then enables access to edit them. However, this problem is highly ambiguous since multiple combinations of intrinsic properties may result in the same image. Many prior work has tackled this ambiguity either by assuming at least one of the intrinsic properties is known [13] or devising priors about each of these properties [1]. The recent success of deep learning based methods, on the other hand, has stimulated research to learn such priors directly from the data. A feasible approach is to predict each of the intrinsic properties independently from an input image, i.e. using individual neural networks to predict normals, material, and lighting. However, the interaction of these properties during the image formation process inspires us to devise a joint prediction framework. An alternative common practice of deep learning methods is to represent the intrinsic properties in a latent feature space [12] where an image can be *decoded* from implicit feature representations. Directly editing such feature representations, however, is not trivial since they do not correspond to any true physical property. On the contrary, the physics of image formation process is well understood and motivates us to replace this *black-box decoder* with a *physical decoder* and thus enables the network to learn the physics of image formation.

We present an end-to-end network architecture that replicates the image formation process in a physically based rendering layer. Our network is composed of prediction modules that infer each of the intrinsic properties from a single image of an object. These predictions are then provided to a rendering layer which re-synthesizes the input image. We define the loss function as a weighted sum of the error over the individual predictions and perceptual error over the synthesized images. We provide comparisons with and without incorporating the perceptual error and show that the combined loss provides significant improvements (see Figure 5 and 6). Due to the lack of a large scale dataset of real images with ground truth normal, material, and lighting annotations, we train our network on a rendering based synthetic dataset. Nevertheless, this network performs reasonable predictions for real images where the space of materials and lighting is much more diverse. We further refine our results on real images with a post-optimization process and show that the network predictions provide a good initialization to this highly non-linear optimization. This paves the road to plausible editing results (see Figure 1).

Our main contributions are:

**1**. We present an end-to-end network architecture for image-based material editing that encapsulates the forward image formation process in a rendering layer.

**2**. We present a differentiable rendering layer that supports both diffuse and specular materials and utilizes an environment map to represent a variety of natural illumination conditions. This layer can trivially be integrated into other networks to enable in-network image synthesis and thus boost performance.

## 2. Related Work

**Intrinsic Image Decomposition.** A closely related problem to image-based material editing is *intrinsic image decomposition* which aims to infer intrinsic properties, e.g. shape, illumination, and material, from a single image and thus enables access to edit them. The work of Barrow et al. [2] is one of the earliest to formulate this problem and since then several variants have been introduced.

Intrinsic image decomposition is an ill-posed problem since different combinations of shape, illumination, and material may result in the same image. Therefore, an important line of work assumes at least one these unknowns to be given, such as geometry. Patow et al. [20] provides a survey of earlier methods proposed to infer illumination, material, or combination of both for scenes with known geometry. More recent work from Lombardi et al. [13, 14] introduces a Bayesian formulation to infer illumination and material properties from a single image captured under natural illumination. They utilize a material representation based on Bidirectional Reflectance Distribution Functions (BRDFs) and thus can handle a wide range of diffuse and specular materials. Approaches that aim to infer all three intrinsic properties [1], on the other hand, often make simplified prior assumptions such as diffuse materials and low-frequency lighting to reduce the complexity of the problem.

An important sub-class of intrinsic image decomposition is *shape from shading (ShS)* where the goal is to reconstruct accurate geometry from shading cues [28]. Many ShS approaches, however, assume prior knowledge about the material properties [9] or coarse geometry [27] to be given.

As opposed to these approaches, we assume no prior knowledge about any of the intrinsic properties to be given and handle both diffuse and specular materials under natural illumination. Instead of making assumptions, we aim to infer priors directly from data via a learning based approach.

**Material Editing.** Some previous methods treat material editing as an image filtering problem without performing explicit intrinsic image decomposition. Khan et al. [10] utilize simple heuristics to infer approximate shape and illumination from an image and utilize this knowledge to perform material editing. Boyadzhiev et al. [3] introduce several image filters to change properties such as shininess and glossiness. While these approaches achieve photo-realistic results they can provide limited editing scenarios without explicit knowledge of the intrinsic properties.

**Material Modeling via Learning.** With the recent success of learning based methods, specifically deep learning, several data-driven solutions have been proposed to infer intrinsic properties from an image. Tang et al. [25] introduce *deep lambertian networks* to infer diffuse material properties, a single point light direction, and an orientation map from a single image. They utilize Gaussian Restricted Boltzmann Machines to model surface albedo and orientation. Richter et al. [22] use random forests to extract surface patches from a database to infer the shape of an object with diffuse and uniform albedo. Narihira et al. [17] predict relative lightness of two image patches by training a classifier on features extracted by deep networks. Similarly, Zhou et al. [29] use a convolutional neural network (CNN) to predict relative material properties of two pixels in an image and then perform a constrained optimization to solve for the albedo of the entire image. Narihira et al. [24] propose a CNN architecture to directly predict albedo and shading from an image. Kulkarni et al. [12] use variational auto-encoders to disentangle viewpoint, illumination, and other intrinsic components (e.g. shape, texture) in a single image. One of the limitation of these methods is the ability to handle diffuse materials only. The recent work of Rematas et al. [21] predict the combination of material and illumination from a single image handling both diffuse and specular materials. In a follow-up work [6], they propose two independent network architectures to further disentangle material and illumination from such a combined representation. However, any error that occurs when inferring the combined representation is automatically propagated to the second step. In contrast, we provide an end-to-end network architecture that performs this disentangling in one pass. Last but not least, several recent work [23, 7] uses neural networks to estimate per-pixel intrinsic properties from a given image. However, these estimations are aligned with the input image, thus, it is not easy transfer these properties across images of different objects as our method does.

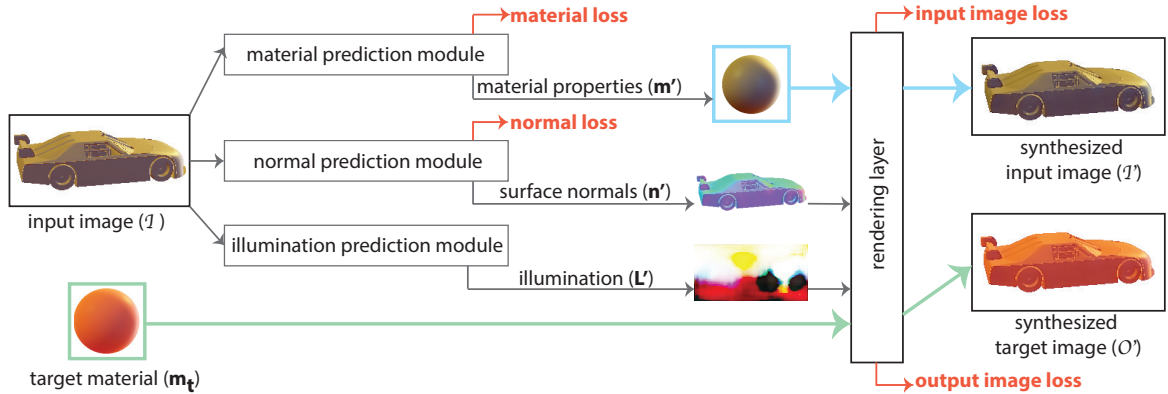

Figure 2. Given a single image of an object, $I$, we propose a network architecture to predict material ($\mathbf{m}'$), surface normals ($\mathbf{n}'$), and illumination ($\mathbf{L}'$). These predictions are provided to a rendering layer which re-synthesizes the input image ($\mathbf{I}'$). In addition, a desired target material, $\mathbf{m_t}$, is passed to the rendering layer along with $\mathbf{n}'$ and $\mathbf{L}'$ to synthesize a target image $\mathcal{O}'$ which depicts the object with the target material from the same viewpoint and under the same illumination. By defining a joint loss that evaluates both the synthesized images and the individual predictions, we perform robust image decomposition and thus enable material editing applications.

## 3. Approach

**Overview.** We present an end-to-end network architecture for image-based material editing. The input is $\mathcal{I}$, a single image of an object $s$ with material $\mathbf{m}$ captured under illumination $\mathbf{L}$. We assume the object is masked out in the image. Given a desired target material definition $\mathbf{m_t}$, the goal is to synthesize an output image $\mathcal{O}$ that depicts $s$ from the same viewpoint with material $\mathbf{m_t}$ and illuminated by $\mathbf{L}$.

While inferring illumination, material, and shape from a single image is an ill-posed problem, once these parameters are known, the forward process of image synthesis, i.e. *rendering* is well defined. We propose a network architecture that encapsulates both this inverse decomposition and the forward rendering processes as shown in Figure 2. The first part of the network aims to infer illumination ($\mathbf{L}'$), material ($\mathbf{m}'$), and 3D shape (represented as surface normals, $\mathbf{n}'$) from $\mathcal{I}$ via three prediction modules. The output of these modules is then provided to a *rendering layer* to synthesize $\mathcal{I}'$. In addition, the target material $\mathbf{m_t}$ is passed to the rendering layer along with predicted $\mathbf{L}'$ and $\mathbf{n}'$ to synthesize an output image $O'$. We define a joint loss that evaluates both the outputs of the rendering layer and the predictions of two of the individual modules. We show that this joint loss significantly improves the accuracy of the inverse decomposition and enables compelling material editing results.

We first introduce our render layer (Sec. 3.1), prediction modules providing data to it (Sec. 3.2), and the training procedure (Sec. 3.3). Then, we discuss how objects composed of multiple materials are handled (Sec. 3.4).

### 3.1. Rendering Layer

The color of each pixel in an image depends on how the corresponding object surface point reflects and emits incoming light along the viewing direction. Our rendering layer aims to replicate this process as accurately as possible. Compared to previously proposed differentiable renderers [15], the main advantage of our rendering layer is to model surface material properties based on BRDFs to handle both diffuse and specular materials. We also utilize environment maps which provide great flexibility in representing a wide range of illumination conditions. We note that our rendering layer also makes several moderate assumptions. We assume the image is formed under translation-invariant natural illumination (i.e. incoming light depends only on the direction). We also assume there is no emission and omit complex light interactions like inter-reflections and subsurface scattering.

Under these conditions, we model the image formation process mathematically similar to Lombardi et al [13]. Given per-pixel surface normals $\mathbf{n}$ (in camera coordinates), material properties $\mathbf{m}$ and illumination $\mathbf{L}$, the outgoing light intensity for each pixel $p$ in image $\mathcal{I}$ can be written as an integration over all incoming light directions $\omega_i$:

$$\mathcal{I}_p(\mathbf{n_p}, \mathbf{m}, \mathbf{L}) = \int f(\omega_i, \omega_o, \mathbf{m}) \mathbf{L}(\omega_i) \max(0, \mathbf{n_p} \cdot \omega_i) d\omega_i, \quad (1)$$

where $\mathbf{L}(\omega_i)$ defines the intensity of the incoming light and $f(\omega_i, \omega_o, \mathbf{m})$ defines how this light is reflected along the outgoing light direction $\omega_o$ based on the material properties $\mathbf{m}$. In order to make this formulation differentiable, we substitute the integral with a sum over a discrete set of incoming light directions defined by the illumination $\mathbf{L}$:

$$\mathcal{I}_p(\mathbf{n_p}, \mathbf{m}, \mathbf{L}) = \sum_{\mathbf{L}} f(\omega_i, \omega_o, \mathbf{m}) \mathbf{L}(\omega_i) \max(0, \mathbf{n_p} \cdot \omega_i) d\omega_i. \quad (2)$$

We now describe how we represent each property and re-fer to the supplementary material about the details of the forward and back propagation on this rendering layer.

**Surface normals (n).** n is represented by a 3-channel image, same size as the input image, where each pixel $p$ encodes the corresponding per-pixel normal $\mathbf{n_p}$.

**Illumination (L).** We represent illumination with an HDR environment map of dimension $64 \times 128$. Each pixel coordinate in this image can be mapped to spherical coordinates and thus corresponds to an incoming light direction $\omega_i$ in Equation 8. The pixel value stores the intensity of the light coming from this direction.

**Material (m).** We define $f(\omega_i, \omega_o, \mathbf{m})$ based on BRDFs [18] which provide a physically correct description of pointwise light reflection both for diffuse and specular surfaces. Non-parametric models [16] aim to capture the full spectrum of BRDFs via lookup tables that encode the ratio of the reflected radiance to the incident radiance given incoming and outgoing light directions $(\omega_i, \omega_o)$. Although such lookup tables achieve highly realistic results, they are computationally expensive to store and not differentiable. Among the various parametric representations, we adopt the Directional Statistics BRDF (DSBRDF) model [19] which is shown to accurately model a wide variety of measured BRDFs. This model represents each BRDF as a combination of hemispherical exponential power distributions and the number of parameters depends on the number of distributions utilized. Our experiments show that utilizing 3 distributions provides accurate approximations resulting in 108 parameters per material definition in total. We refer the reader to the original work [19] for more details.

## 3.2. Prediction Modules

We utilize three prediction modules to infer surface normals, material properties, and illumination. The input to each module is the $256 \times 256$ input image $\mathcal{I}$. We refer to the supplementary material for the detailed network architecture of each module.

**Normal prediction.** The normal prediction module follows the same spirit as the recent work of Eigen et al. [5]. The main difference is that we predict a normal map that is equal in size to the input image by utilizing a 4-scale-submodule network as opposed to the originally proposed 3-scale-submodule network. The fourth submodule consists of 2 convolutional layers where both input and output size is equal to the size of the input image. We utilize a normalization layer to predict surface normals with unit length.

**Illumination prediction.** Illumination prediction module is composed of 7 convolutional layers where the output of each such layer is half the size of its input. The convolutional layers are followed by 2 fully connected layers and a sequence of deconvolutional layers to generate an environment map of size $64 \times 128$. Each convolutional and fully connected layer is accompanied by a rectifier linear unit. Fully connected layers are also followed by dropout layers.

**Material prediction.** Material prediction module is composed of 7 convolutional layers where the output of each such layer is half the size of its input. The convolutional layers are followed by 3 fully connected layers and a tanh layer. Each convolutional and fully connected layer is accompanied by a rectifier linear unit. Fully connected layers are also followed by dropout layers. We note that since each of the 108 material parameters are defined at different scales, we normalize each to the range $[-0.95, 0.95]$ and remap to their original scales after prediction.

## 3.3. Training.

We train the proposed network architecture by defining a joint loss function that evaluates both the predictions of the individual modules and the images synthesized by the rendering layer. Specifically, we define a L2 normal loss:

$$l_{normal} = \sum_p (\mathbf{n_p} - \mathbf{n'_p})^2, \qquad (3)$$

where $\mathbf{n_p}$ and $\mathbf{n'_p}$ denote the ground-truth and predicted surface normals for each pixel $p$ respectively. Since both ground-truth and predicted normals are unit length, this loss is equivalent to cosine similarity used by Eigen et al. [5].

We define the material loss as the L2 norm between the ground truth (**m**) and predicted material parameters (**m'**):

$$l_{material} = (\mathbf{m} - \mathbf{m'})^2 \qquad (4)$$

We also utilize a perceptual loss, $l_{perceptual}$, to evaluate the difference between the synthesized $\{\mathcal{I}', \mathcal{O}'\}$ and ground truth $\{\mathcal{I}, \mathcal{O}\}$ images, which helps to recover normal details not captured by the L2 normal loss. We use the pre-trained *vgg16* network to measure $l_{perceptual}$ as proposed by Johnson et al. [8].

We define the final loss, $l$, as a weighted combination of these individual loss functions:

$$l = w_n l_{normal} + w_m l_{material} + w_p l_{perceptual}, \quad (5)$$

where we empirically set $w_n = 1 \times 10^4$, $w_m = 1 \times 10^3$, $w_p = 1$.

Note that we do not define a loss between ground truth and predicted illumination due to the limited amount of available public HDR environment maps for training. Each pixel in the spherical environment map corresponds to a light direction. Depending on the viewing direction and surface normals, the contribution of each light direction to the final rendered image will vary. We leave it as future work to design a robust loss function that takes this non-uniformity into account.

## 3.4. Extension to Multi-Material Case

A typical use case scenario for image-based material editing is where a user denotes the region of interest of an object and a desired target material that should be applied to this region. Our approach trivially supports such use cases. Given a segmentation mask denoting regions of uniform material, we perform material prediction for each region while predicting a single normal and illumination map for the entire image. All these predictions are then provided to the rendering layer. Target materials are defined for each region of interest separately. The rest of the training and testing process remains unchanged.

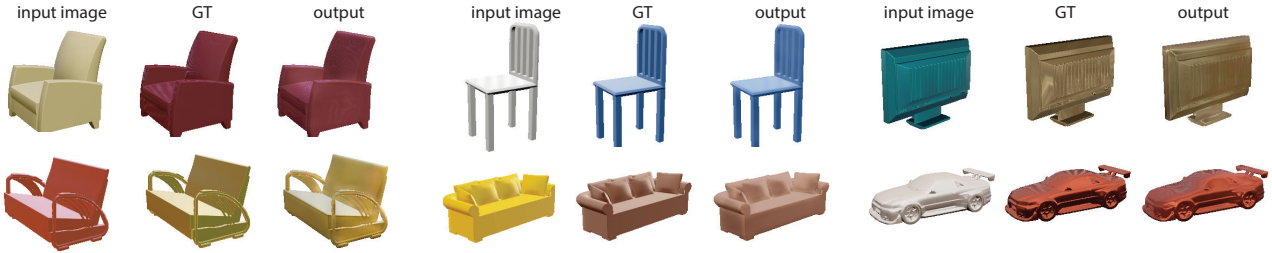

Figure 3. For each input image, we synthesize an output image with a given target material definition. We provide the ground truth (GT) target images for reference. The left (middle) column shows cases where the target material is more (less) specular than the input material. We also provide examples where both the input and the target material are specular in the right column.

### 3.5. Refinement with Post-optimization

Due to the lack of a large scale dataset of real images with ground truth normal, material, and illumination annotations, we train our network with synthetic renderings. Although large 3D shape repositories [4] provide sufficient shape variation, the network does not see a large spectrum of material and illumination properties captured in real images. While increasing the variation in the synthetic data could reduce the discrepancy between the training and test sets, the gap is almost always there theoretically. Thus, we refine our network predictions on real images with a post-optimization. Specifically, using $\mathbf{n}', \mathbf{m}', \mathbf{L}'$, the network predictions of surface normal, material, and illumination for an input image $\mathcal{I}$ as initialization, we optimize for $\mathbf{n}^*, \mathbf{m}^*, \mathbf{L}^*$ which minimize the following energy function:

$$\underset{\mathbf{n}^*,\mathbf{m}^*,\mathbf{L}^*}{\mathrm{argmin}} \ \| \mathcal{I}^* - \mathcal{I} \|^2 + a \| \mathbf{n}^* - \mathbf{n}' \|^2 + b \| \mathbf{L}^* - \mathbf{L}' \|^2 .$$
(6)

$\mathcal{I}^*$ is the image formed by $\mathbf{n}^*, \mathbf{m}^*, \mathbf{L}^*$ and the first term penalizes the image reconstruction loss. The remaining regularization terms penalize the difference between the network predictions of normal and illumination and their optimized values. We experimentally set $a=1$, $b=10$ as the relative weighting of these regularization terms. We use L-BFGS to solve Equation 6 in an alternative scheme where only one intrinsic property is updated at each iteration.

## 4. Evaluation

In this section we provide quantitative and qualitative evaluations of our method and comparisons to other work.

### 4.1. Datasets and Training

We train the framework with a large amount of synthetic data generated for *car, chair, sofa*, and *monitor* categories. We utilize a total of 280 3D models (130 cars, 50 chairs, 50 sofas and 50 monitors) obtained from ShapeNet [4]. For materials, we use BRDF measurements corresponding to 80 different materials provided in the MERL database [16]. For illumination, we download 10 free HDR environment maps from the Internet and use random rotations to augment them.

**Data generation.** For each 3D model, material, and illumination combination we render 5 random views from a fixed set of pre-sampled 24 viewpoints around the object with fixed elevation. We split the data such that no shape and material is shared between training and test sets. Specifically, we use 80 cars for pretraining, and 40 shapes per each category for joint category finetuning, which leaves 10 shapes per category for testing. Out of the 80 materials, we leave 20 for testing and use the rest in pre-training and training. This split allows us to generate $240K$ pre-training instances, and $480K$ multi-category finetuning instances. In total, the network is trained with over $720K$ unique material-shape-light configurations.

**Training procedure.** We initialize the weights from a uniform distribution: $[-0.08, 0.08]$. Normal and material modules are pre-trained using L2 loss for a few iterations and then trained jointly with illumination network. We utilize Adam [11] optimizer using stochastic gradient descent. Similarly, we first used momentum parameters $\beta_1 = 0.9$ and $\beta_2 = 0.999$, and a learning rate of $0.0001$. Later we reduced the learning rate to be $1 \times 10^{-6}$ and $1 \times 10^{-8}$. We also reduced $\beta_1$ to 0.5 for a more stable training.

### 4.2. Evaluation on Synthetic Data

To test our network, we randomly select an image corresponding to shape $s$, environment map $\mathbf{L}$, and material $\mathbf{m}$ from our test set as input. Given a target material $\mathbf{m_t}$, an output image depicting $s$ with material $\mathbf{m_t}$ and illuminated by $\mathbf{L}$ is synthesized. We compare this synthesized output to ground truth using two metrics. While the L2 met-ric measures the average pixel-wise error, SSIM metric[26] measures the structural difference. We compute the L2 error on tone mapped images where each color channel is in the range [0, 255] and define per-pixel error as the average of the squared difference of three color channels. We note that lower (higher) numbers are better for the L2 (SSIM) metric. We provide quantitative results in Table 3, last two columns. We note that our network is trained for all categories jointly. Figure 3 provides visual results demonstrating that our approach can successfully synthesize specularities when replacing the input material with a more specular target material. Similarly, specularities in the input image are successfully removed when replacing the input material with a more diffuse one.

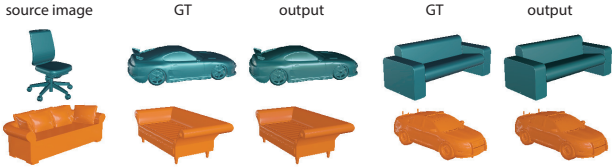

Figure 4. We synthesize an image of an object under some desired illumination with material properties predicted from a source image. We provide the ground truth (GT) for reference.

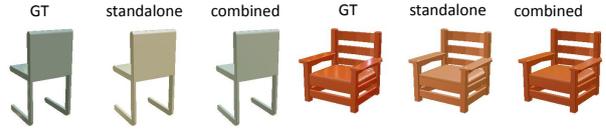

Figure 5. For each example we show the ground truth image as well as the renderings obtained by utilizing the material coefficients predicted by the standalone material prediction module and the combined approach which also uses the rendering layer.

In Figure 4, we show examples where material properties predicted from a source image are transferred to a different object under some desired illumination. Such material transfers alleviates the need to define target materials explicitly, instead each image becomes an exemplar.

### 4.3. Evaluation of the Rendering Layer

We evaluate the effectiveness of utilizing the perceptual loss on the images synthesized by the rendering layer as opposed to independent predictions. We note that for this evaluation, we directly use the pure network predictions without post-optimization.

**Accuracy of Material Coefficients.** For this evaluation, given an input image $\mathcal{I}$, we provide ground truth normals (**n**) and illumination (**L**) and use our network to predict only material. We also train the material prediction module as a standalone network. While both the standalone module and our network use the L2 loss on material coefficients in spirit similar to the recent work of Georgoulis et al. [6], the latter combines this with the perceptual loss on $\mathcal{I}'$. Note that during training, the material parameters have been normalized to the range [-0.95,0.95] to reduce the scale issues for L2 loss; while in Table 1, we remap to the original scales. 60 materials are used for training and 20 for testing. We provide qualitative and quantitative results in Figure 5 and Table 1 respectively. The results demonstrate that the L2 loss is not sufficient in capturing the physical material properties, a small L2 error may result in big visual discrepancies in the rendering output. Incorporation of the rendering layer resolves this problem by treating the material coefficients in accordance with their true physical meaning.

|  | material | rendering | |
|---|---|---|---|
|  | L2 | L2 | SSIM |
| standalone | 4.1038 | 544.1 | 0.9297 |
| combined | 4.8144 | 355.5 | 0.9517 |

Table 1. We show the accuracy of the predicted material coefficients and the images rendered using these together with ground truth normals and illumination. We provide results of the material prediction module trained standalone vs with the rendering layer.

**Accuracy of Surface Normals.** For this evaluation, given an input image $\mathcal{I}$, we provide ground truth material (**m**) and illumination (**L**) and use our network to predict only surface normals. We use the *car* category for the evaluation. The predicted surface normals along with **m** and **L** are used to synthesize $\mathcal{I}'$. We also train the normal prediction module as a standalone network. While both networks use the L2 loss on the surface normals, our network combines this with the perceptual loss on $\mathcal{I}'$.

We provide qualitative and quantitative results in Figure 6 and Table 2 respectively. We observe that L2 loss helps to get the orientation of the surface normals correct, and thus results in small errors, but produces blurry output. Adding the perceptual loss helps to preserve sharp features resulting in visually more plausible results. We also note that, although commonly used, L2 and SSIM metrics are not fully correlated with human perception. Incorporation of the rendering layer results in slim improvements in quantitative numbers (Table 1 and 2) but significantly better appearance results as shown in Figure 5 and 6.

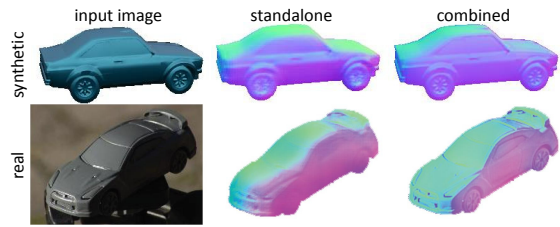

Figure 6. For a synthetic and a real input image, we provide the surface normals predicted by the normal prediction module when trained standalone vs with the rendering layer.

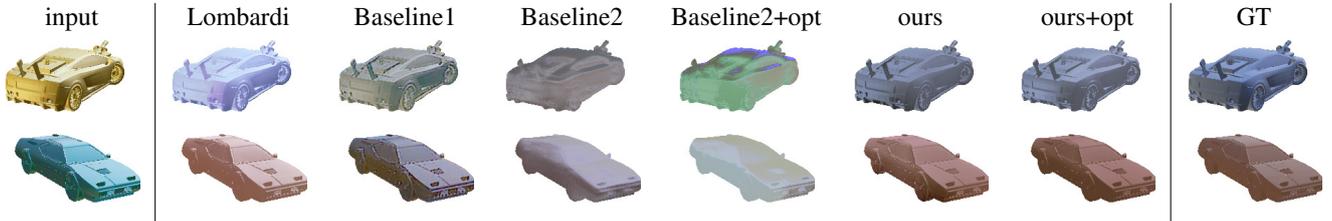

Figure 7. For each example we show the input and the ground truth target image (GT). We provide the results obtained by the method of Lombardi et al. [13], two baseline methods, and our approach.

|            | normals | rendering |        |
|------------|---------|-----------|--------|
|            | cosine  | L2        | SSIM   |
| standalone | 0.9105  | 313.2058  | 0.9441 |
| combined   | 0.9050  | 238.4619  | 0.9625 |

Table 2. We show the accuracy of the predicted surface normals and the images rendered using these together with ground truth material and illumination. We provide the results of the normal prediction module trained standalone vs with the rendering layer.

### 4.4. Multi-Material Examples

We show examples for objects composed of multiple materials. For these examples, we manually segment 100 chair models to two semantic regions where each region gets its own material assignment. We generate renderings from 5 different viewpoints with random illumination and materials assignments. We fine-tune the model trained on single-material examples with these additional renderings. We utilize 80 chairs for fine-tuning and 20 for testing. In Figure 8, we show examples of editing each region with a target material. We also quantitatively evaluate these results with respect to ground truth over 4000 examples. We achieve an L2 error of 1272.1 and SSIM index of 0.9362. No post-optimization is used for the experiments.

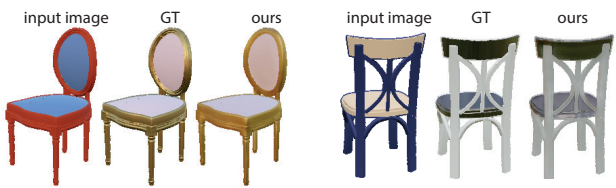

Figure 8. Given an image of a multi-material object, we show how different target material assignments are realized for different parts. We provide ground target (GT) images for reference.

### 4.5. Comparisons

We compare our method with several baseline methods and previous approaches which we briefly discuss below and refer to the supplementary material for more details.

**Lombardi et al. [13]** We compare our method to the approach of Lombardi et al. [13] which uses certain priors to optimize for illumination and material properties in a single image of an object with known geometry. We provide the ground truth surface normals to this method and predict the material and illumination coefficients. We render the target image given the target material, ground truth normals, and the predicted illumination.

**Baseline 1.** Our first baseline is inspired by the work of Kulkarni et al. [12] and is based on an encoder-decoder network. The encoder maps the input image to a feature representation, $f = (f_m, f_o)$, with one part corresponding to the material properties ($f_m$) and the remaining part corresponding to other factors ($f_o$) (i.e. normals and illumination). This disentangling is ensured by passing $f_m$ as input to a regression network which is trained to predict the input material coefficients. $f_m$ is then replaced by the feature representation of the target material. The decoder utilizes the updated feature representation to synthesize the target image. We also provide skip connections from the input image to the decoder to avoid blurry output. Similar to our approach, *baseline 1* is trained with the perceptual loss between the synthesized and ground truth target image and the L2 loss on the estimated input material coefficients.

**Baseline 2.** For the second baseline, given the input image we train three individual networks to predict the surface normals, material, and illumination separately. All networks are trained with the L2 loss. We then render the target image using the predicted surface normals, illumination, and the target material.

For this comparison we train our method and both of the baselines on the same training and testing set. We refine the output of baseline 2 and our network with the post-optimization described in Section 3.5. We use 500 images sampled from the testing set as input and generate a new image with a target material from the test set. We provide quantitative (see Table 3) and visual results (see Figure 7). For baseline 2, the use of L2 loss only results in blurry normal prediction, hence the blurry target images. While baseline 1 addresses this issue by use of skip connections, implicit feature representations fail to capture the true intrinsic properties and result in a large appearance difference with respect to ground truth. The approach of Lombardi et al. [13] performs non-convex optimization over the ren-dering equation which is likely to get stuck in local minimum without a good initialization. The individual predictions from baseline 2 are quite far from the global minimum or any reasonable local minimum. Thus, we observe no improvement with post-optimization, in some cases the optimization in fact gets stuck in a worse local minimum. Our initial network output performs in par with [13] with no assumption on known surface normals. Refining these predictions with the post-optimization outperforms all other methods significantly.

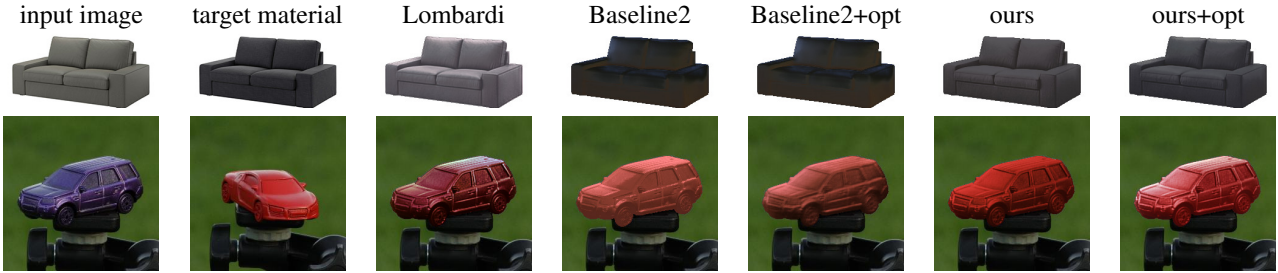

Figure 9. We transfer the target material to the input image using the method of Lombardi et al. [13], baseline 2, baseline 2 with post-optimization, our network, and our network with post-optimization.

|      | Lombardi | Baseline1 | Baseline2 | | ours | |
|------|----------|-----------|-----------|-----------|-----------|-----------|
|      |          |           | no opt | opt | no opt | opt |
| L2   | 802.7    | 1076.7    | 886.9 | 1273.7 | 805.2 | 530.9 |
| SSIM | 0.9416   | 0.9173    | 0.9256 | 0.9159 | 0.9408 | 0.9557 |

Table 3. We evaluate the accuracy of material transfer results on synthetic data for Lombardi et al. [13], baseline 1, baseline 2 with and without post-optimization, and our method with and without post-optimization.

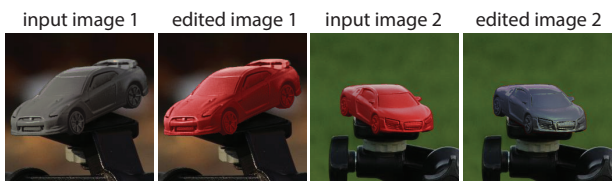

Figure 10. Given two input images, we synthesize new images by swapping the material properties of the objects.

### 4.6. Evaluation on Real Images

We provide visual comparisons of our method with the method of Lombardi et al. [13] and baseline 2 on real product images[1] downloaded from the internet in Figure 9. Since the method of Lombardi et al. [13] assumes surface normals to be known, we provide the normal prediction of our network to their optimization. Additionally, we refine the results of both baseline 2 (i.e. individual prediction of normal, material, and illumination) and our network with the post-optimization of Section 3.5. Even though our network has been trained only on synthetic data with a limited set of environment maps and materials, the raw network results are promising and provide a good initialization to the post-optimization. Independent predictions, on the other hand, result in the post-optimization to get stuck in a local minimum which does not yield as visually plausible results.

---

[1]For real examples we finetune our network using all materials in our dataset to better generalize to unseen materials.

Finally, in Figure 10, we provide material transfer examples on some real images provided by the SMASHINg challenge dataset [21] with the combination of our network and the post optimization. We refer to the supplementary material for more examples and comparisons.

## 5. Conclusion and Future Work

We propose an end-to-end network architecture for image-based material editing that encapsulates the image formation process in a rendering layer. We demonstrate various plausible material editing results both for synthetic and real data. One of the limitations of our work is the fact that lighting space is learned with a relatively small dataset, because of the lack of data. This may result in imperfect decomposition, specifically some lighting effects being predicted to be part of the surface normals. For multi-material cases, estimating material for a small segment independently is difficult. For such scenarios, extending the network to perform per-pixel material predictions will be a promising direction. Incorporation of advanced light interactions such as subsurface scattering and inter-reflections in the rendering layer is crucial to generalize the method to scenes with multiple objects and real images. We expect the performance boost obtained by the encapsulation of the rendering layer to stimulate further research in designing neural networks that can replicate physical processes.

**Acknowledgement**

We would like to thank Xin Sun, Weilun Sun, Eunbyung Park, Chunyuan Li and Liwen Hu for helpful discussions. This work is supported by a gift from Adobe, NSF EFRI-1240459, CNS-1205260 and AFOSR FA9550-17-1-0075.

# Appendix

## Details of Rendering Layer

We model the image formation process mathematically similar to the classical rendering equation, but without emitted spectral radiance. Given per-pixel surface normals $\mathbf{n}$ (in camera coordinates), material properties $\mathbf{m}$ and illumination $\mathbf{L}$, the outgoing light intensity for each pixel $p$ in image $\mathcal{I}$ can be written as an integration over all incoming light directions $\omega_i$:

$$\mathcal{I}_p(\mathbf{n_p}, \mathbf{m}, \mathbf{L}) = \int f(\overrightarrow{\omega}_i, \overrightarrow{\omega}_p, \mathbf{m}) \mathbf{L}(i) \max(0, \mathbf{n_p} \cdot \overrightarrow{\omega}_i) d\omega_i, \quad (7)$$

where $\mathbf{L}(\omega_i)$ defines the intensity of the incoming light and $f(\omega_i, \omega_p, \mathbf{m})$ defines how this light is reflected along the outgoing light direction $\omega_p$ based on the material properties $\mathbf{m}$. $\omega_p$ is also the viewing direction, which can be computed using $FOV$ and image size. In order to make this formulation differentiable, we substitute the integral with a sum over a discrete set of incoming light directions defined by the illumination $\mathbf{L}$:

$$\mathcal{I}_p(\mathbf{n_p}, \mathbf{m}, \mathbf{L}) = \sum_{\mathbf{L}} f(\overrightarrow{\omega}_i, \overrightarrow{\omega}_p, \mathbf{m}) \mathbf{L}(i) \max(0, \mathbf{n_p} \cdot \overrightarrow{\omega}_i) d\omega_i. \quad (8)$$

where $d\omega_i$ represents the contribution (weight) a single light $\omega_i$.

## Representations

We now describe in detail how surface normals, illumination, and material properties are represented.

**Surface normals (n).** Given an image $\mathcal{I}$ of dimension $w \times h$, $\mathbf{n}$ is represented by a 3-channel $w \times h$ normal map where the $r$, $g$, $b$ color of each pixel $p$ encodes the $x$, $y$, and $z$ dimensions of the per-pixel normal $\overrightarrow{n_p}$. The normal for each pixel has 3 channels:

$$\mathbf{n_p} = (n_p^1, n_p^2, n_p^3)$$

**Illumination (L).** We represent illumination with an HDR environment map of dimension $64 \times 128$. This environment map is a spherical panorama image flattened to the 2D image domain. Each pixel coordinate in this image can easily be mapped to spherical coordinates and thus corresponds to an incoming light direction $\omega_i$ in Equation 8. The pixel value stores the intensity of the light coming from this direction.

Let $H_L$ and $W_L$ represent the height and width of the environment map respectively. For each pixel $i = h * W_L + w$, which has the row index and column index to be $h_L$ and $w_L$, in the environment map, we define $\theta_i^L$ and $\phi_i^L$ to be:

$$\theta_i^L = \frac{h_L}{H_L}\pi, \quad \phi_i^L = \frac{w_L}{W_L}\pi$$

Then the direction of the lighting this pixel generates is:

$$\overrightarrow{\omega_i} = <\cos\phi_i^L \sin\theta_i^L, \cos\theta_i^L, \sin\phi_i^L \sin\theta_i^L>$$

Note that we will not compute the derivative of $\overrightarrow{\omega_i}$ and there is no parameter to learn during the training.

**Material (m).** We define $f(\omega_i, \omega_p, \mathbf{m})$ based on BRDFs [18] which provide a physically correct description of pointwise light reflection both for diffuse and specular surfaces. We adopt the Directional Statistics BRDF (DSBRDF) model [19] which is shown to accurately model a wide variety of measured BRDFs. The DSBRDF is based on approximating the BRDF values using the mixtures of hemi-sphere distributions.

To begin with, we define a *half vector* to be:

$$\overrightarrow{h_p} = \frac{\overrightarrow{\omega_i} + \overrightarrow{\omega_p}}{||\overrightarrow{\omega_i} + \overrightarrow{\omega_p}||}$$

We then denote the angle between half vector and lighting direction to be $\theta_d$.

$$\theta_d = acos(min(1, max(0, \overrightarrow{\omega_i} \cdot \overrightarrow{h_p})))$$

The material coefficient is related to $\theta_d$. For each $\theta_d$, the material coefficient has 3 ($<R, G, B>$ channels) $\times$ 3 (3 mixtures of hemi-sphere distribution) $\times$ 2 (2 coefficients per hemi-sphere distribution) parameters. Instead of tabulating the material coefficients for every $\theta_d$, we only estimate the coefficients of a few $\theta_d$-s. Specifically, 18 $\theta_d$-s' corresponding coefficients are estimated using the raw MERL BRDF dataset [16]. Later, those coefficients will be used to fit a second degree B-spline with nine knots, which results in 6 variables [13]. Thus, in total, there will be $3 \times 3 \times 2 \times 6$ parameters in each BRDFS's material coefficient $\mathbf{m}$.

We denote the second degree B-spline function as $S(\theta_d, \mathbf{m})$, then for any $\theta_d$, the coefficients are $(m_{s,t}^k) = S(\theta_d, \mathbf{m})$, where $k \in \{0, 1, 2\}$, $s \in \{0, 1, 2\}$ and $t \in \{0, 1\}$. $k$ represents one of the 3 channels (R, G, B); $s$ represents one of the three mixtures of hemi-sphere distributions; $t$ indexes two coefficients in hemi-sphere distribution.

The function $f(\overrightarrow{\omega}_i, \overrightarrow{\omega}_p, \mathbf{m})$ can be re-written as:

$$f^k(\overrightarrow{\omega}_i, \overrightarrow{\omega}_p, \mathbf{m}) = \sum_{s=0}^{2} (e^{m_{s,0}^k \cdot max(0, \overrightarrow{h}_p \cdot \mathbf{n_p})^{m_{s,1}^k}} - 1)$$

where $k$ represents one of the 3 channels (R, G, B).

Having all of these representations, we can re-write our image formation as following:

$$\mathcal{I}_p^k(\mathbf{n_p}, \mathbf{m}, \mathbf{L}) = \sum_{i=1}^{H_L \cdot W_L} f^k(\overrightarrow{\omega}_i, \overrightarrow{\omega}_p, \mathbf{m}) \mathbf{L}(i) \max(0, \mathbf{n_p} \cdot \overrightarrow{\omega}_i) d\omega_i$$

$$= \sum_{i=1}^{H_L \cdot W_L} (\sum_{s=0}^{2} (e^{m_{s,0}^k \cdot max(0, \overrightarrow{h}_p \cdot \mathbf{n_p})^{m_{s,1}^k}} - 1)) \cdot L^k(i) \cdot \max(0, \mathbf{n_p} \cdot \overrightarrow{\omega}_i) \sin(\frac{\lfloor i/W_L \rfloor}{H_L} \pi)$$

**Derivative**

**Derivative over Light.** For lighting, we only need to compute the derivative over the intensity values of $L^k(i), k \in \{0, 1, 2\}$. We don't need to compute the derivative of the lighting direction.

$$\frac{\partial \mathcal{I}_p(\mathbf{n_p}, \mathbf{m}, \mathbf{L})}{\partial L^k(i)} = (\sum_{s=0}^{2} (e^{m_{s,0}^k \cdot max(0, \overrightarrow{h}_p \cdot \mathbf{n_p})^{m_{s,1}^k}} - 1)) \cdot \max(0, \mathbf{n_p} \cdot \overrightarrow{\omega}_i) \sin(\frac{\lfloor i/W_L \rfloor}{H_L} \pi)$$

**Derivative over Normal.** We compute the derivative of normal for each channel individually. If $\overrightarrow{h}_p \cdot \mathbf{n_p} <= 0$ or $\mathbf{n_p} \cdot \overrightarrow{\omega}_i <= 0$, then $\frac{\partial \mathcal{I}_p(\mathbf{n_p}, \mathbf{m}, \mathbf{L})}{\partial n_p^c} = 0$

Otherwise,

$$\frac{\partial \mathcal{I}_p(\mathbf{n_p}, \mathbf{m}, \mathbf{L})}{\partial n_p^c} = \sum_{k=0}^{2} \sum_{i=1}^{H_L \cdot W_L} (\sum_{s=0}^{2} (e^{m_{s,0}^k \cdot max(0, \overrightarrow{h}_p \cdot \mathbf{n_p})^{m_{s,1}^k}} - 1)) \cdot L^k(i) \cdot \omega_i^c \sin(\frac{\lfloor i/W_L \rfloor}{H_L} \pi) +$$

$$\sum_{k=0}^{2} \sum_{i=1}^{H_L \cdot W_L} (\sum_{s=0}^{2} (e^{m_{s,0}^k \cdot max(0, \overrightarrow{h}_p \cdot \mathbf{n_p})^{m_{s,1}^k}} \cdot m_{s,0}^k \cdot m_{s,1}^k \cdot max(0, \overrightarrow{h}_p \cdot \mathbf{n_p})^{m_{s,1}^k - 1}) \cdot$$

$$L^k(i) \cdot \max(0, \mathbf{n_p} \cdot \overrightarrow{\omega}_i) \sin(\frac{\lfloor i/W_L \rfloor}{H_L} \pi)$$

**Derivative over Material.** We first compute the derivative for $m_{s,t}^k$, and then based on chain rule and the spline interpolation function, we get the derivative for the original $\mathbf{m}$.

$$\frac{\partial \mathcal{I}_p(\mathbf{n_p}, \mathbf{m}, \mathbf{L})}{\partial m_{s,0}^k} = \sum_{i=1}^{H_L \cdot W_L} (e^{m_{s,0}^k \cdot max(0, \overrightarrow{h}_p \cdot \mathbf{n_p})^{m_{s,1}^k}} \cdot max(0, \overrightarrow{h}_p \cdot \mathbf{n_p})^{m_{s,1}^k}) \cdot L^k(i) \cdot \max(0, \mathbf{n_p} \cdot \overrightarrow{\omega}_i) \sin(\frac{\lfloor i/W_L \rfloor}{H_L} \pi)$$

$$\frac{\partial \mathcal{I}_p(\mathbf{n_p}, \mathbf{m}, \mathbf{L})}{\partial m_{s,1}^k} = \sum_{i=1}^{H_L \cdot W_L} (e^{m_{s,0}^k \cdot max(0, \overrightarrow{h}_p \cdot \mathbf{n_p})^{m_{s,1}^k}} \cdot m_{s,0}^k \cdot max(0, \overrightarrow{h}_p \cdot \mathbf{n_p})^{m_{s,1}^k} \cdot ln(\overrightarrow{h}_p \cdot \mathbf{n_p})) \cdot$$

$$L^k(i) \cdot \max(0, \mathbf{n_p} \cdot \overrightarrow{\omega}_i) \sin(\frac{\lfloor i/W_L \rfloor}{H_L} \pi)$$

then applying back the spline interpolation function $(m_{s,t}^k) = S(\theta_d, \mathbf{m})$, we have
$\frac{\partial \mathcal{I}_p(\mathbf{n_P}, \mathbf{m}, \mathbf{L})}{\partial \mathbf{m}} = \left(\frac{\partial \mathcal{I}_p(\mathbf{n_P}, \mathbf{m}, \mathbf{L})}{\partial m_{s,t}^k}\right) \cdot \frac{\partial S(\theta_d, \mathbf{m})}{\partial \mathbf{m}}$

**Network Design**

We provide the detailed architectures for the normal, material, and illumination prediction modules in Figures 11, 12, and 13 respectively. Those networks can also be replaced with some other more recent network designs. The goal of this paper is to show performance improvement brought by the rendering layer.

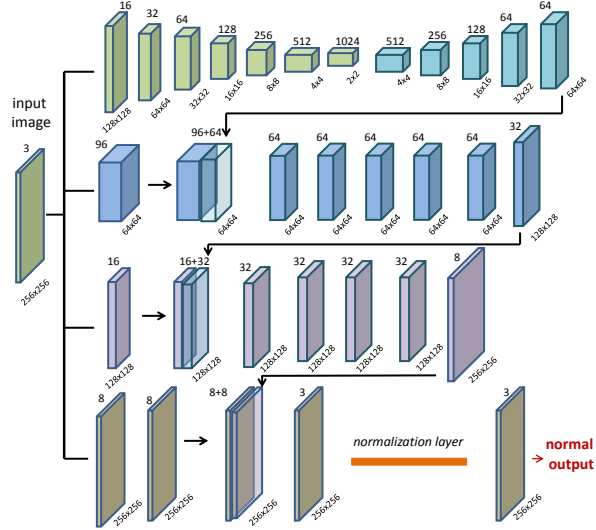

Figure 11. Architecture for the normal prediction module

**Comparison with Deep Reflectance Map**

We provide evaluations of our approach on the real images provided in the SMASHINg Challenge dataset provided by Rematas et al. [21]. For these examples, we train our model on all the 80 material definitions in our dataset to better generalize to unseen materials and viewpoints in the real images. We provide visual results in Figures 14. Specifically, we provide visualizations of reflectance maps defined as the orientation-dependent appearance of a fixed material under a fixed illumination by Rematas et al. [21]. In addition to the reflectance maps predicted by our network, the reflectance maps with post-optimization using our network prediction as initial guess, and the various method presented by Rematas et al. [21] we also provide the visualization of the material coefficients (under a fixed natural light), illumination, and normals from the post-optimization using our network predictions as initial guess.

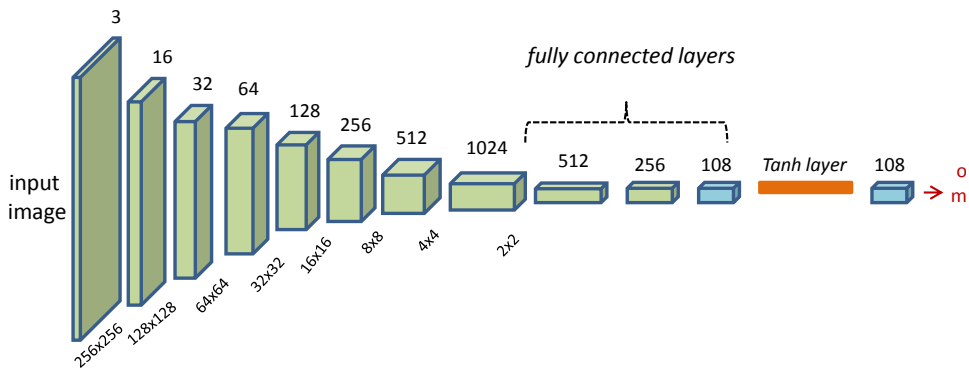

Figure 12. Architecture for the material prediction module

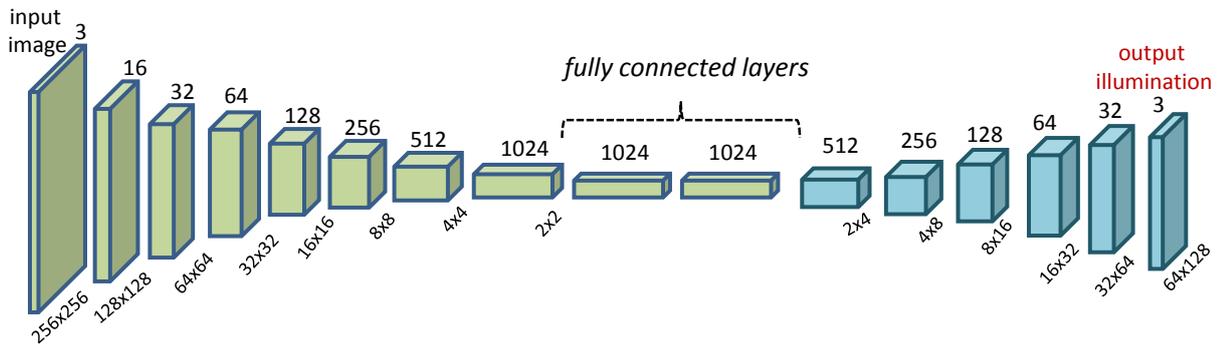

Figure 13. Architecture for the illumination prediction module

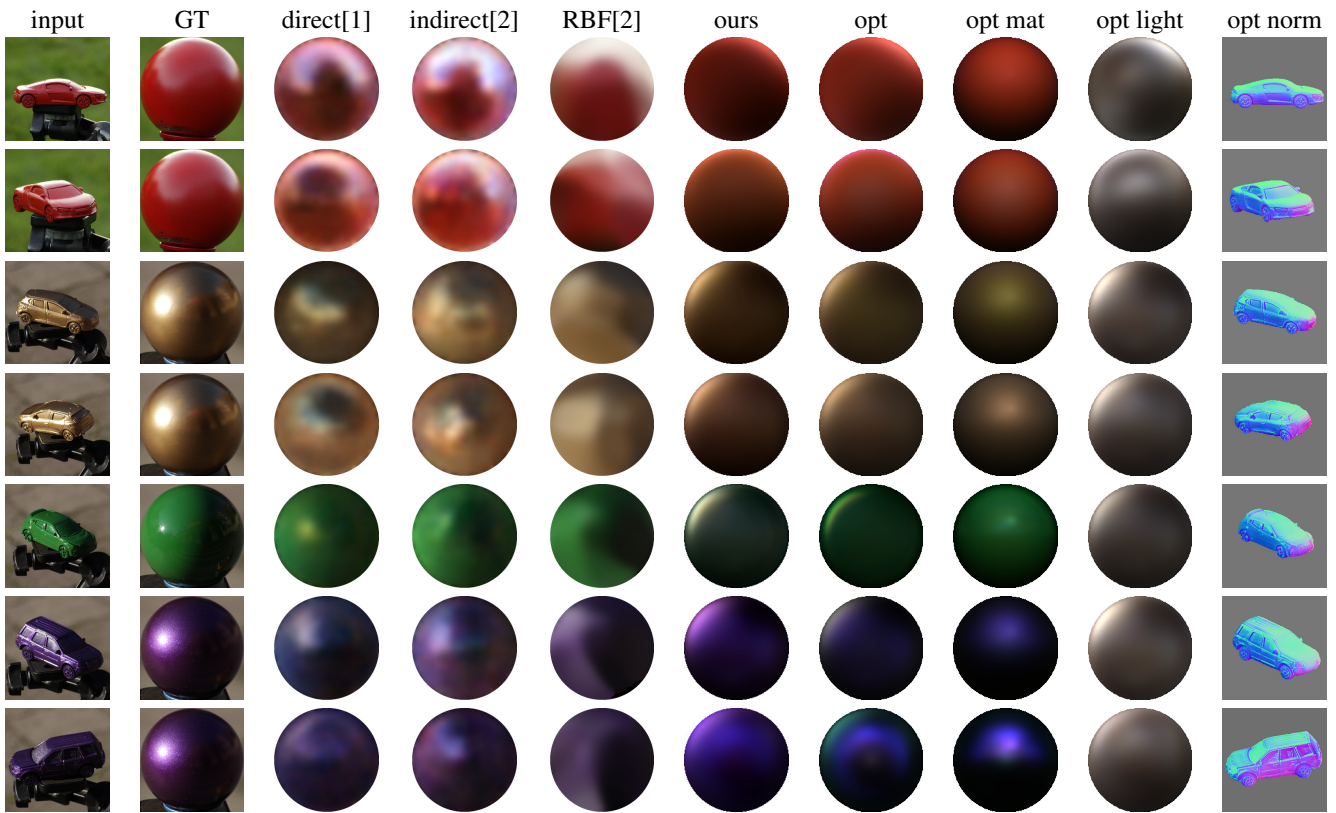

Figure 14. For each input image, we provide ground truth (GT) reflectance maps as well as those predicted by the various methods presented by Rematas et al. [21]: direct & indirect & indirect with RBF interpolation. For our approach, we provide the initial network predicted reflectance maps, post-optimized reflectance maps using network prediction as initialization, as well as individual components of post-optimized material, illumination, and normals.

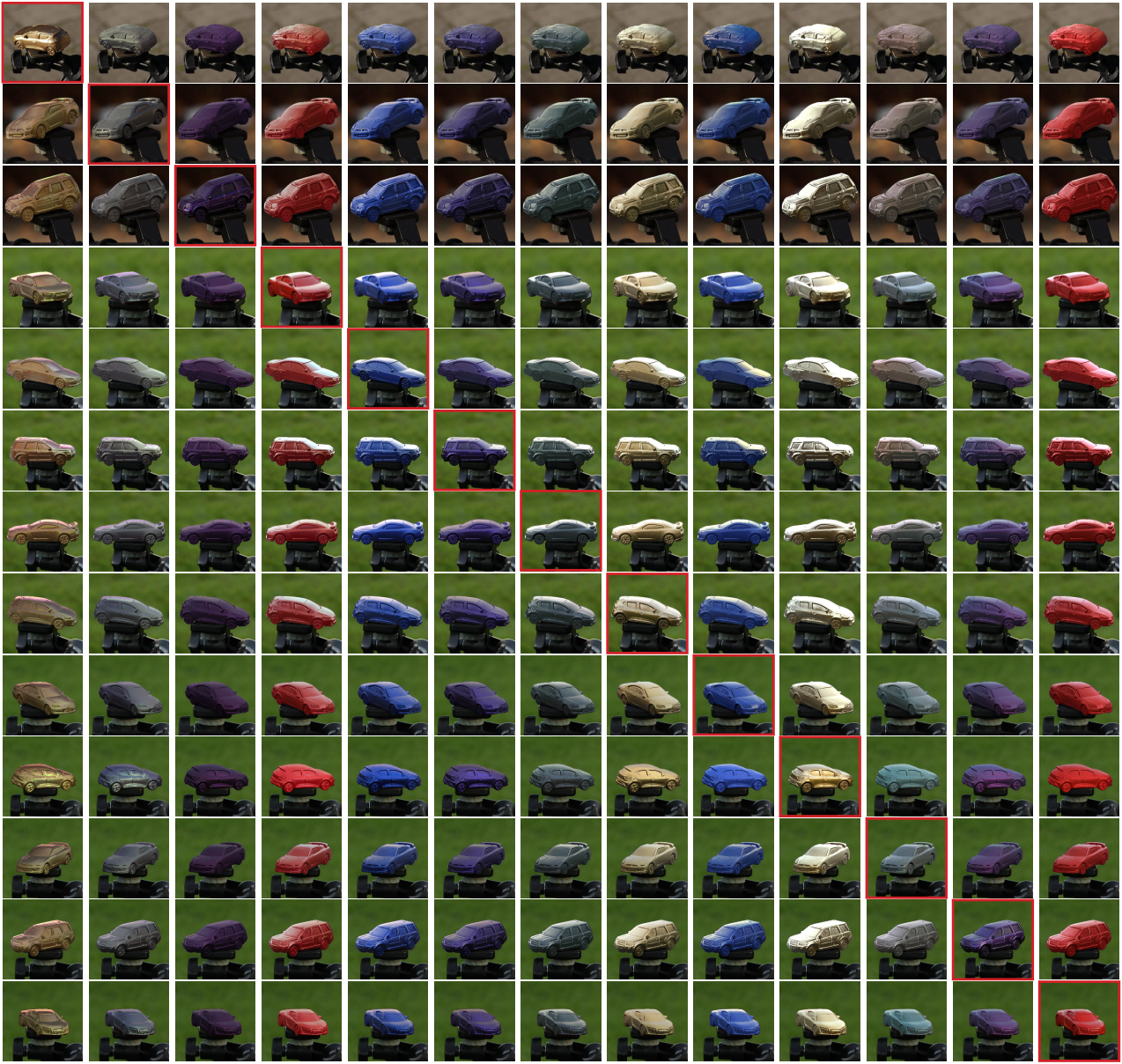

Figure 15. Given a set of images (in diagonal, in red boxes) as inputs, we synthesize new images by using shape and light from its row and material from its column using our approach.

## Cross Material Transfer

In Fig 15, we provide the cross material transferring results, where we transfer the material predicted from one image to another image, using our approach with post-optimization.